\documentclass[final]{hldlike}

\usepackage{amsfonts}
\usepackage{booktabs}
\usepackage{microtype}
\usepackage{nicefrac}
\usepackage{mathtools}
\usepackage{hyperref}
\usepackage{enumitem}
\graphicspath{{figures/}}

\newtheorem{corr}{Corollary}

\newtheorem{proposition}{Proposition}
\usepackage{algorithm}
\usepackage{algpseudocode}
\usepackage{bm}

\title[CMRL: Interference and Rich-Learning]{Center-Manifold Reduction of Learning at Bifurcations: Interference and Rich Learning in Recurrent Neural Networks}

\hldauthor{\Name{James Hazelden} \Email{jhazelde@uw.edu}\\
        \addr{Applied Mathematics, University of Washington\\
        The Allen Institute}
        \AND
        \Name{Eric Shea-Brown} \Email{etsb@uw.edu}\\
        \addr{Applied Mathematics and Neurobiology \& Biophysics, University of Washington\\
        The Allen Institute}
}

\newcommand{\R}{\mathbb{R}}
\newcommand{\NTK}{\mathrm{GeNTK}}
\newcommand{\FIM}{\mathrm{FIM}}

\newcommand{\norm}[1]{\left\lVert #1 \right\rVert}
\newcommand{\inner}[2]{\left\langle #1, #2 \right\rangle}

\newcommand{\op}{\mathrm}
\newcommand{\eqn}[2]{\begin{align} \label{eqn:#1} #2 \end{align}}

\begin{document}
\maketitle
\thispagestyle{empty}
\pagestyle{plain}

\vspace{-0.6cm}

\begin{abstract}
Rich learning in recurrent neural networks often proceeds through sudden transitions in latent dynamics, but there is little theory predicting how gradient descent behaves during these events. We study the local learning geometry near codimension-one bifurcations through the global empirical Neural Tangent Kernel (GeNTK). Under local center-manifold conditions, and when bifurcation-related sensitivity dominates bounded residual terms, we show that the global parameter-to-state Jacobian \(D_\theta h\) is approximated by a low-rank normal-form operator. The induced GeNTK and Fisher information matrix therefore become strongly amplified and anisotropic, concentrating toward a rank-one channel for the four scalar codimension-one bifurcations and a rank-two real channel for a Neimark--Sacker bifurcation. Controlled high-dimensional RNN experiments validate this operator reduction. In learned RNNs, the same low-rank concentration coincides with abrupt loss changes and subtask interference, while a local projection predicts the sign of these effects near isolated events. Finally, in an input-driven 15-task LeakyRNN, GeNTK amplification aligns with continuation-detected changes in the MemoryPro dynamics. These results suggest a tractable operator-level description of learning near dynamical transitions, together with scalable diagnostics for amplified low-dimensional learning geometry.
\end{abstract}


\section{Introduction}
\begin{figure}[t]
    \centering
    \includegraphics[width=\linewidth]{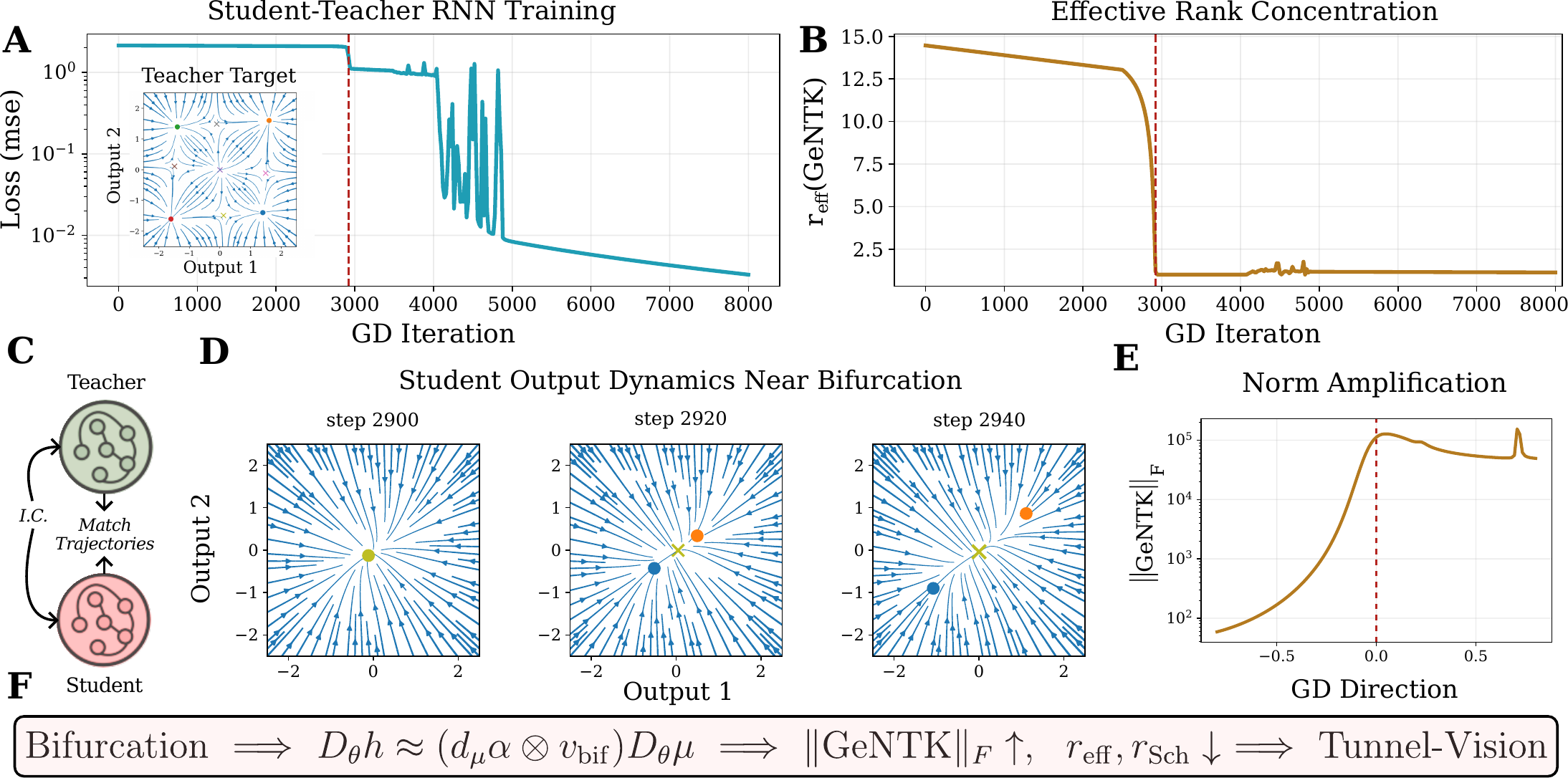}
    \caption{\vspace{-1.0em} \textbf{A pitchfork bifurcation during student-teacher RNN training and induced GeNTK operator distortion. }
    \textbf{A} Student-teacher RNN training exhibits a sharp loss drop near a learned bifurcation (see Appendix~\ref{app:student_teacher} for details).
    \textbf{B} At the same transition, the GeNTK effective rank collapses toward one.
    \textbf{C} The student is trained to match teacher trajectories from shared initial conditions.
    \textbf{D} Output-space snapshots show the emergence of a supercritical pitchfork.
    \textbf{E} The GeNTK norm is sharply amplified along the local GD direction, consistent with concentration onto a dominant bifurcation channel.
    \textbf{F} Schematic of Theorem~\ref{thm:cmfd} as well as Corollaries~\ref{corr:sig} and~\ref{cor:bif_interference}.}
    \label{fig:rnn}
\end{figure}

Gradient descent (GD) in dynamical systems can be perilous. Training frequently encounters long plateaus, exploding or vanishing gradients, and interference between subtasks. Sometimes GD steers a model into a rich regime, where its hidden dynamics organize into interpretable motifs that support generalization. Other times, it simply overfits in a largely lazy way, reusing dynamics already present near initialization. What determines these outcomes remains poorly understood: the evolution of latent dynamics under GD is generally intractable, with recent theory still largely restricted to cases such as linear or stationary dynamics~\citep{saxe2013exact,lukas2024from,bordelon2025dynamicallylearningintegraterecurrent,clark2025connectivity}.

In the rich regime, solving a task may require creating, destroying, or reorganizing fixed points, limit cycles, and other dynamical motifs. Classical dynamical systems theory tells us that such qualitative changes occur through local bifurcations~\citep{guckenheimer1983nonlinear}: sudden nonlinear events in which a fixed point may split, change stability, or give rise to an orbit. Here, we study the five canonical codimension-one bifurcations of discrete dynamical systems.

Prior work has shown that bifurcations can coincide with abrupt drops in loss during recurrent-network training~\citep{eisenmann2023bifurcations}. But why should a change in the hidden dynamics cause a sudden change in learning? Which task components benefit, which are delayed or forgotten, and what does GD actually do near the bifurcation boundary? We study these questions through the \textit{global empirical Neural Tangent Kernel} (\(\NTK\)), the Gram operator
\[
\NTK=(D_\theta h)(D_\theta h)^T
\]
of the global parameter-to-state Jacobian, where the state consists of empirical samples over time and a batch of task inputs. Recent tools make this operator interpretable and computationally accessible in finite recurrent models~\citep{hazelden2026globalempiricalntkselfreferential}, providing a direct window into the geometry of learning near bifurcations.
Note that \(D_\theta h\) is crucial to GD learning and second-order algorithms. Specifically, all error signals filter through this operator to produce parameter changes. \(\NTK\) describes the first-order state-space effect of these corrections, but it can also be seen simply as the Gram matrix induced by the exact operator \(D_\theta h\), through which parameter changes exactly pass, so that its eigenvalues describe the spectrum of this operator, i.e., the local landscape and bias of GD at a particular point. 

Our central finding is that, near an isolated bifurcation and under the conditions made explicit below, learning can become strongly amplified and effectively low-dimensional. At a codimension-one transition, a single tensor-valued mode of \(D_\theta h\) can dominate, or a two-dimensional real channel for a complex conjugate pair, causing the induced \(\NTK\) to concentrate toward a low-rank operator. GD is then funneled through a narrow set of dynamical corrections. Task components aligned with this channel can be learned rapidly, while orthogonal or opposing components may be delayed or worsen locally. We call this \textit{bifurcation-induced tunnel vision}. In this regime, large gradients near bifurcations have a structured geometry that can reorganize the latent representation and produce sudden changes in total and subtask losses.

To make this precise, we develop \textit{Center-Manifold Reduction of Learning} (CMRL), extending classical normal-form reduction from the dynamics themselves to the parameter-to-state learning operator. CMRL reduces the high-dimensional Jacobian \(D_\theta h\) to a low-rank, separable operator determined by the corresponding normal form, with explicit conditions controlling the approximation error. The resulting operator predicts the amplified state-space direction, the local GD flow, and the interaction between task components. Put simply, CMRL lets us study rich representational transitions in high-dimensional models using classical, analytically tractable normal forms.

This structure also suggests a possible intervention. Natural-gradient methods can recondition stiff learning geometry, but exact natural gradients are usually infeasible in recurrent systems, while inexpensive adaptive methods need not recover the relevant operator geometry~\citep{martens2020new,marschall2020unified,kunstner2019limitations}. Because CMRL identifies the amplified component as low-rank, we also explore a low-rank natural-gradient correction as a secondary test of whether reconditioning a small critical subspace can stabilize training near these transitions.

To summarize, our \textbf{contributions} are:
\begin{enumerate}
    \item We develop \textit{Center-Manifold Reduction of Learning} (CMRL), an operator-level approximation of \(D_\theta h\) by a low-rank, separable normal-form channel. We give explicit conditions under which the approximation error becomes small and derive corresponding low-rank structure in the GeNTK and Fisher information matrix.
    \item We validate the operator prediction in controlled dense-coupled 32-dimensional tanh RNNs containing each of the five canonical codimension-one bifurcations. The full parameter-to-trajectory operator concentrates toward the continuation-derived critical subspace as the event is approached.
    \item We derive a local \textit{bifurcation-induced tunnel vision} prediction: when one amplified channel dominates, the sign of each subtask loss change depends on its alignment with the same state-space field. In a joint fixed-point--orbit task, this prediction captures the signs of the main loss changes near isolated events and coincides with low-rank latent reorganization.
    \item We explore two extensions: a low-rank natural-gradient correction that stabilizes the joint task when a small number of modes are reconditioned, and matrix-free GeNTK diagnostics in an input-driven 15-task LeakyRNN, where amplification aligns with continuation-detected changes in MemoryPro dynamics.
\end{enumerate}

\section{Related Work}

\paragraph{Rich learning and recurrent dynamics.}
Lazy learning reuses features or dynamics already present near initialization, whereas rich learning substantially reorganizes the learned representation~\citep{chizat2019lazy,liu2024how,farrell2023from,chakravarthula2026task}. In recurrent networks, task solutions often organize around low-dimensional attractors, structured connectivity, and dynamical motifs that may be reused across tasks or created during training~\citep{NEURIPS2020sch,turner2023simplicity,driscoll2024flexible,bordelon2025dynamicallylearningintegraterecurrent,clark2025connectivity,marschall2025multitask,clark2026structure}. We study the latter event: the local geometry of GD when a new motif is formed through a bifurcation.

\paragraph{Bifurcations, criticality, and information geometry.}
Classical center-manifold and normal-form theory reduce local bifurcation dynamics to canonical low-dimensional systems~\citep{guckenheimer1983nonlinear}. Near a loss of stability, critical slowing down produces large long-time sensitivities and ill-conditioned parameter inference~\citep{roesch2019parameter}. Related information-geometric and sloppy-model analyses identify singular Fisher geometry and stiff parameter combinations near bifurcations~\citep{dasilva2022information,anderson2023sloppy}, while bifurcations during RNN training have been linked directly to abrupt changes in loss~\citep{eisenmann2023bifurcations}. Our target is the empirical parameter-to-state operator \(D_\theta h\) itself. We show when, near a codimension-one transition,
\[
D_\theta h
\approx
(d_\mu \alpha \otimes v_{\mathrm{bif}})D_\theta\mu,
\]
retaining the spatiotemporal state direction and finite-horizon error signal required to predict local GD updates. The induced operators
\[
\NTK := (D_\theta h)(D_\theta h)^\ast,
\qquad
\FIM := (D_\theta h)^\ast(D_\theta h),
\]
then inherit the same amplified, low-rank structure.

\paragraph{The NTK as a diagnostic of learning.}
Recent work has reinforced the importance of studying tractable local events during training which can be studied analytically~\citep{simon2026scientific}.
The NTK was introduced to characterize function-space learning in the infinite-width limit~\citep{jacot2018neural}, with extensions to recurrent, equilibrium, and general architectures~\citep{alemohammad2020recurrent,yang2020tensorprogramsII,feng2023deqntk}. Its spectrum and alignment have since been used to study learning rates, spectral bias, generalization, feature learning, and finite-width evolution~\citep{bordelon2020spectrum,shan2022theoryneuraltangentkernel,baratin2021implicitregularizationneuralfeature,canatar2022outofdistributiongeneralizationkernelregression,bordelon2023dynamics,bordelon2024feature,pmlr-v139-yang21c,Lee_2020}. Here, we use the finite, time-varying global empirical NTK (\(\NTK\), or GeNTK) as a diagnostic of rich learning rather than as a fixed-kernel approximation. Its operator factorization exposes how recurrent response dynamics filter parameter geometry~\citep{hazelden2026globalempiricalntkselfreferential}. Randomized matrix-free methods further make its amplification, effective rank, alignment, and Schmidt rank accessible without forming the full operator~\citep{novak2022fastfinitewidthneural,mohamadi2023fast,hazelden2025fastneuraltangentkernel}. In this view, peaks in \(\|\NTK\|\), drops in effective rank, and drops in Schmidt rank provide scalable signatures of sudden rich transitions, including input-dependent transitions for which classical autonomous bifurcation analysis is less direct.

\paragraph{Interference, forgetting, and multitask computation.}
Bifurcation-driven reorganization of attractors and their basins has recently been linked directly to memory formation, spurious memories, and catastrophic forgetting in Hebbian Hopfield networks~\citep{essex2025memorisation}. Multitask RNNs may likewise reuse shared attractors and dynamical motifs, select among task-dependent computations, or isolate task structure to reduce interference~\citep{lea2020,turner2023simplicity,driscoll2024flexible,marschall2025multitask}. Our contribution is complementary: rather than tracking only the creation and destruction of attractors, we identify the local GD geometry that can produce interference and forgetting. Near a bifurcation, \(\NTK\) concentrates onto a small number of amplified channels, forcing otherwise distinct subtask errors through nearly the same state-space direction. Aligned components improve together, while orthogonal or opposing components slow down, interfere, or worsen. This gives an explicit operator-level account of the tunnel vision described in Corollary~\ref{cor:bif_interference}.

\section{Center-Manifold Reduction of Learning}

Here, we present the central theoretical framework: approximating the operators relevant to first-order learning geometry by a simpler form characterized by normal forms at codimension-one bifurcations. 
This reduction, when error bounds are satisfied (as in Theorem~\ref{thm:cmfd}), results in a set of three testable criteria characterizing the behavior of the learning operators at the bifurcation, as well as an explicit low-rank approximation to the full operator.
The consequences (sudden changes in loss and subtask interference) are explored further in Section~\ref{sec:consequences} and our experiments. 
First, we present the approximation (Algorithm~\ref{alg:cmrl}), then quantify the error bounds (Theorem~\ref{thm:cmfd}), with the projection crucially relying on amplification of bifurcation-relevant learning geometry. 
In the canonical normal forms, this amplification follows from the slow decay of perturbations as the critical multiplier approaches the unit circle, consistent with classical critical-slowing-down and information-geometric analyses~\citep{guckenheimer1983nonlinear,roesch2019parameter,dasilva2022information,anderson2023sloppy}. 
Figure~\ref{fig:valthm} provides a controlled numerical validation of
Theorem~\ref{thm:cmfd}, showing that the predicted low-rank operator becomes
increasingly accurate as each bifurcation is approached.

\subsection{Algorithm}

\paragraph{Classical CMR Approximation} Classically, center manifold reduction (CMR) is a procedure for approximating dynamics local to a bifurcation.
Specifically, for a codimension-one bifurcation at a fixed point (FP) \(\bar h\) of a discrete dynamical system of the form 
\eqn{dyn}{h_{t+1} = f(h_t, \theta)}
with state \(h_t \in \R^N\) and parameters \(\theta \in \R^P\), CMR first defines local coordinates near the bifurcation, 
\eqn{cmrparam}{\theta \leftrightarrow (\mu, R)}
with \(\mu \in \R\) regulating the bifurcation (occurring at \(\mu = 0\)) and \(R \in \R^{P-1}\) the residual center-manifold parameters. 
From this, CMR can be used to rewrite the dynamics in the form 
\eqn{cmrstate}{h = \bar h + \alpha(\mu) \otimes v + \varepsilon(\mu, R, \alpha)}
where \(\alpha\) is a low-dimensional \textit{normal form} with dynamics quantified by one of five explicit dynamical systems (Table~\ref{table:nforms}), \(v\in \R^N\) is the state-space direction relevant to the bifurcation and \(\varepsilon\) captures higher-order terms and residual dynamics. Crucially, the error \(\varepsilon\) can be bounded assuming sufficient closeness of \(\mu\) to \(0\) and the sample trajectory to \(\bar h\) (made explicit in Theorem~\ref{thm:cmfd}). 

\paragraph{Empirical Learning Operators} CMR is primarily concerned with reducing the model dynamics local to a bifurcation.
However, when the model is the consequence of a learning procedure, the key quantity of interest is how bifurcations affect this learning process and geometry, not only the dynamics themselves. 
Specifically, when training the model in Equation~\ref{eqn:dyn}, the state is typically sampled for a range of \(B\) initial conditions, simulated for \(T\) timesteps each, resulting in an empirical 3-tensor \textit{global state} \(h \in \R^{B \times T \times N}\) (input-dependent tasks are explored in Appendix~\ref{app:input_gru}).
To this, one assigns a loss \(L(h) \in \R\) and standard Gradient Descent (GD) perturbs \(\theta\) by \(\delta \theta = -\nabla_\theta L(h)\).
By the chain rule, it is clear that the crucial linear operator regulating this update is the tensor-valued Jacobian \(D_\theta h\), which we call the \textit{empirical global-state sensitivity}. Specifically,
$$\delta \theta = - \nabla_\theta L = - (D_\theta h)^T(\nabla_h  L)$$
This object defines the first-order landscape, i.e., how the empirical state samples depend on the parameters, and its induced Gram matrices, the \textit{Neural Tangent Kernel} and \textit{Fisher Information}, are crucial to understanding learning, inductive bias, and the design of higher-order optimizers~\citep{jacot2018neural,martens2020new,kunstner2019limitations,karakida2020understanding}.  
Notably, \(D_\theta h\), although potentially massive, is often low-rank and highly structured with clean factorizations, due to the dependence of future states on prior states as well as structured low-rank latent geometry~\citep{hazelden2026globalempiricalntkselfreferential,mastrogiuseppe2018linking,pellegrino2023low}. 
Thus, a generic procedure for CMR-based approximation of the empirical operator \(D_\theta h\) local to bifurcations is desirable to quantify how such events warp the learning geometry.

\paragraph{Extending CMR to Learning Operators} The dynamical approximation in Equation~\ref{eqn:cmrstate} suggests a natural approximation to \(D_\theta h\) by the rank-one, spacetime-separable operator \(d_\mu \alpha \otimes v\), as summarized in Algorithm~\ref{alg:cmrl}. There, \(\text{CMR}\) denotes a generic function taking the state and parameters and returning the relevant terms for the state and parameter projections. Practically, we introduce numerical continuation software in PyTorch and implement this procedure automatically, as detailed in Appendix~\ref{app:continuation}.
Note that the derivation of this operator approximation is a simple application of the chain rule at this empirical, state over time and samples, level. 
However, CMR does not guarantee that this operator approximation must also be valid. 
This instead hinges on Proposition~\ref{prop:amp}, which shows how the bifurcation-related sensitivity can become large relative to the residual terms. 

\begin{algorithm}[H]
\caption{Center-Manifold Reduction of Learning}
\label{alg:cmrl}
\begin{algorithmic}[1]
\Require Codimension-one bifurcation. Empirical state \(h \in \R^{B \times T \times N}\), parameters \(\theta \in \R^P\)
\State \((\mu, R), (\bar h, \alpha \otimes v, \varepsilon) \gets \text{CMR}(\theta, h)\) \hfill  \textcolor{gray}{// Classical dynamics projection, Equations~\ref{eqn:cmrparam} \& ~\ref{eqn:cmrstate}}
\State $\op J_{\text{CMRL}} \gets (d_\mu \alpha \otimes v) \cdot (D_\theta \mu)$\hfill \textcolor{gray}{// Low-rank operator approximation}
\State $\op J_{\text{err}} \gets D_\theta h - \op J_{\text{CMRL}}$
\State \textbf{Return} \(\op J_{\text{CMRL}}, \op J_{\text{err}}\) 
\end{algorithmic}
\end{algorithm}

\paragraph{Observations} Since \(\alpha\), the normal form, has an explicit expression based on the bifurcation type (five cases in Table~\ref{table:nforms}), \(d_\mu \alpha\) can be quantified directly. The key additional condition is that this bifurcation-related sensitivity becomes large relative to the residual sensitivities. When that occurs, it can dominate the high-dimensional operator \(D_\theta h\), producing the low-rank, anisotropic geometry quantified in Theorem~\ref{thm:cmfd} and tested in Figure~\ref{fig:valthm}. 

\paragraph{Efficient Matrix-Free Extraction.}
The same low-rank structure can be probed without explicitly forming the full operator. The \(\NTK\) norm and effective rank can be estimated by stochastic trace estimation~\citep{hazelden2025fastneuraltangentkernel,meyer2021hutchoptimalstochastictrace}, and a dominant field can be recovered by matrix-free power iteration or randomized SVD using Jacobian--vector and vector--Jacobian products. These quantities are best interpreted as diagnostics of amplified low-dimensional learning geometry: a peak or rank collapse need not uniquely identify an exact classical bifurcation, but can indicate a regime in which a small dynamical channel dominates learning.
\subsection{Amplification of Low-Rank Learning Geometry in Normal Forms}

\paragraph{Normal forms} A key ingredient in the accuracy of the approximation to \(D_\theta h\) present in Algorithm~\ref{alg:cmrl} is dominance of the bifurcation-related sensitivity \(d_\mu \alpha\). 
Such dominance occurs when \(d_\mu \alpha\) dramatically amplifies in norm as \(\mu\) passes through \(0\), inducing the relevant codimension-one bifurcation. 
Conveniently, by the CMR construction, \(\alpha(\mu)\) is exactly predicted by a codimension-one normal form with a scalar parameter and low-dimensional state, i.e., of the form 
$$\alpha_{t+1} = g(\alpha_t, \mu)$$
where there are only five distinct cases for \(g\), as in the following table.

\begin{figure}[t]
    \centering
    \includegraphics[width=.98\linewidth]{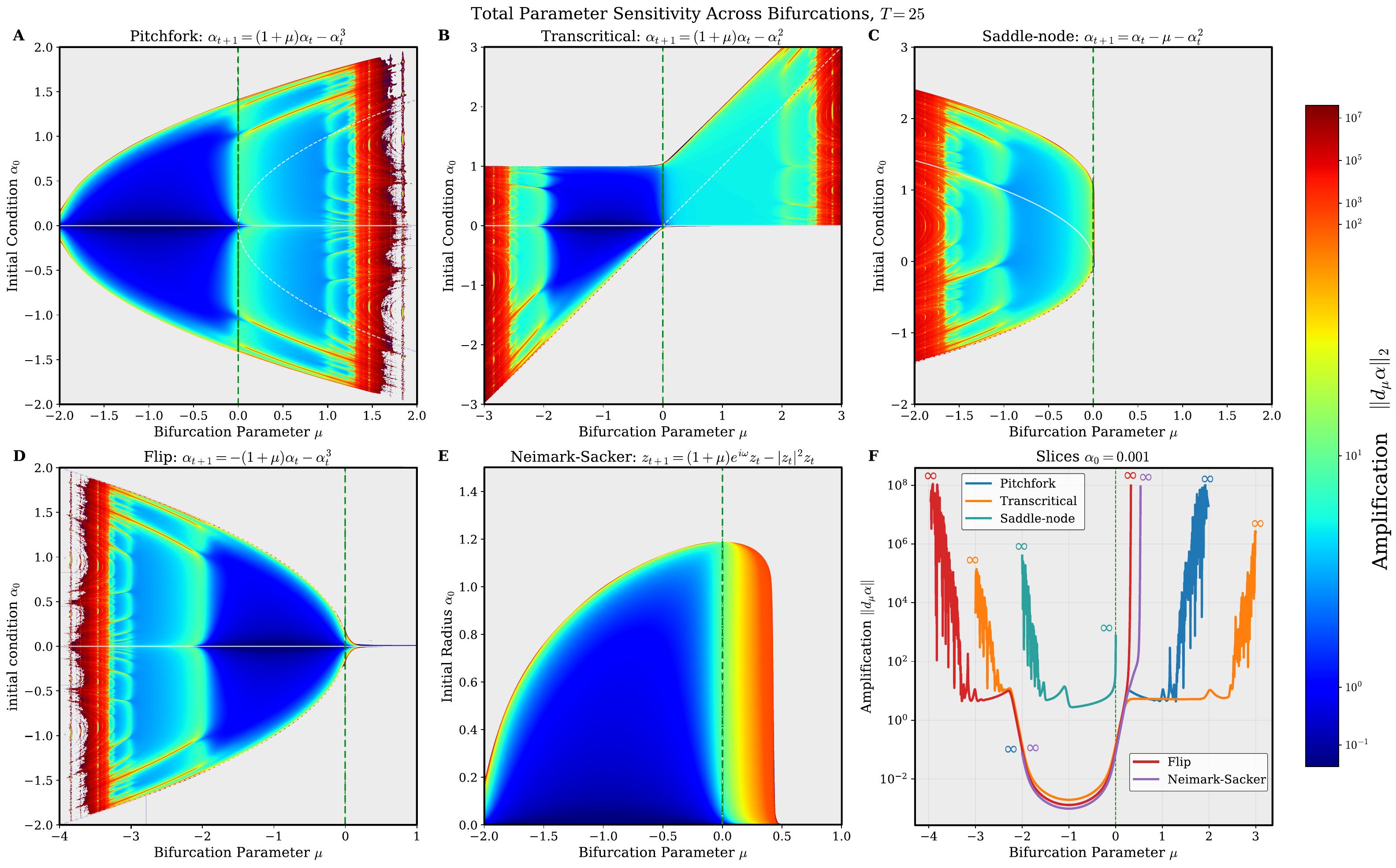}
    \caption{For each normal form, the state is \(\alpha\) and the parameter is \(\mu\), with a bifurcation at \(\mu = 0\) (see Table~\ref{table:nforms}). 
For each bifurcation type, we plot the norm \(\|d_\mu \alpha\|\), where \(\alpha \in \R^T\) is the state of the model, unrolled over \(T = 25\) timesteps, with initial condition \(\alpha_0\) (y-axis) and parameter \(\mu\) (x-axis). 
Gray areas correspond to amplification above \(10^7\), where learning yields severe exploding gradients.
The green line denotes the bifurcation boundary \(\mu =0\), and white lines are fixed-point branches.
The bottom-right panel plots the sample slice at \(0.001\) for the five cases, showing the amplification near \(\mu = 0\).
Curves were cut off when the amplification was \(> 10^7\) (marked with the infinity symbol).}
    \label{fig:normformamp}
\end{figure}

\begin{table}[h]
\centering
\caption{Representative discrete-time codimension-one bifurcation normal
forms. Signs are chosen so that the new stable branches or orbit appear for
\(\mu>0\).}
\label{table:nforms}
\begin{tabular}{||c | c||}
 \hline
 \textbf{Type} & \textbf{RHS,} \(\bm{g}\) \\
 \hline
 Pitchfork (PF)
    & \((1+\mu)h_t-h_t^3\) \\
 \hline
 Transcritical (TC)
    & \((1+\mu)h_t-h_t^2\) \\
 \hline
 Saddle-node (SN)
    & \(h_t+\mu-h_t^2\) \\
 \hline
 Flip (FL)
    & \(-(1+\mu)h_t+h_t^3\) \\
 \hline
 Neimark--Sacker (NS)
    & \(R_\omega\!\left((1+\mu)h_t-\|h_t\|^2h_t\right)\),
      \quad \(h_t\in\mathbb{R}^2\) \\
 \hline
\end{tabular}
\end{table}

Here \(R_\omega\) denotes rotation in the plane through a fixed angle
\(\omega\),
\[
R_\omega=
\begin{pmatrix}
\cos\omega & -\sin\omega\\
\sin\omega & \cos\omega
\end{pmatrix}.
\]
Thus, in the Neimark--Sacker case, \(\mu\) changes the modulus of the
complex-conjugate multiplier pair while \(\omega\) fixes its phase. 
In this case, the bifurcation-induced low-rankness of \(D_\theta h\) is rank two, corresponding to the formation of a two-dimensional orbit, instead of rank one in the other cases.

The explicit normal forms also make it easy to see why
\(d_\mu\alpha\) can become large near a bifurcation.
As the critical multiplier approaches the stability boundary,
perturbations decay more slowly and their effects accumulate over time.
This is closely related to classical critical slowing down
\citep{kramer1985stabilization}. Here, we track its consequence for the
finite-horizon parameter sensitivity \(d_\mu\alpha\), which directly enters
the CMRL approximation.

\begin{proposition}[Finite-Horizon Amplification Near a Bifurcation]
\label{prop:amp}
For
\[
\alpha_{t+1}=g(\alpha_t,\mu),
\]
the sensitivity satisfies
\[
(d_\mu\alpha)_{t+1}
=
\partial_\alpha g(\alpha_t,\mu)(d_\mu\alpha)_t
+
\partial_\mu g(\alpha_t,\mu).
\]
Near a bifurcation, the critical multiplier approaches the unit circle, so
these perturbations decay slowly and accumulate over time. 
At a real critical multiplier \(+1\), with nonzero bounded parameter
forcing, this gives
\[
(d_\mu\alpha)_t=O(t),
\qquad
\|d_\mu\alpha\|_2=O(T^{3/2}).
\]
For the linearized normal form these scale as
\(O(|\alpha_0|t)\) and \(O(|\alpha_0|T^{3/2})\), respectively (except for saddle-node, which drops the \(|\alpha_0|\) term).
Thus, the finite-horizon amplification depends on both the sampled time
window and the trajectory initial condition.
\end{proposition}

\paragraph{Numerical Validation} Figure~\ref{fig:normformamp} shows how \(d_\mu \alpha\) depends on the initial condition \(\alpha_0\) and proximity to the bifurcation. Across the five normal forms, the sensitivity is strongly amplified near \(\mu=0\), with the detailed landscape depending on the initial condition and bifurcation type. In some regions the norm exceeds the plotting threshold of \(10^7\). These sharp changes in scale illustrate how even simple normal forms can produce exploding or vanishing gradients near a bifurcation. 

\subsection{Quantifying CMRL Operator Accuracy}

\paragraph{Key Conditions} We now quantify the relative error
\(\frac{\|\op J_{err}\|_F}{\|\op J_{CMRL}\|_F}\) of the operator approximation
of the CMRL in Algorithm~\ref{alg:cmrl}.
Specifically, the error depends on (1) locality assumptions relevant to the
classical dynamical CMR (Equation~\ref{eqn:cmrstate}) and (2) critical
amplification of \(d_\mu\alpha\) relative to the bounded residual Jacobian
terms.
The first condition essentially says that bifurcations become less relevant
to learning when they are far from the relevant state-space trajectories or
occurred a while ago, as expected.
The second condition follows from the prior section's normal-form analysis.
This is formalized below, with a proof in
Appendix~\ref{app:cmrl_proof}.

\begin{figure}[t]
    \centering
    \includegraphics[width=.9\linewidth]{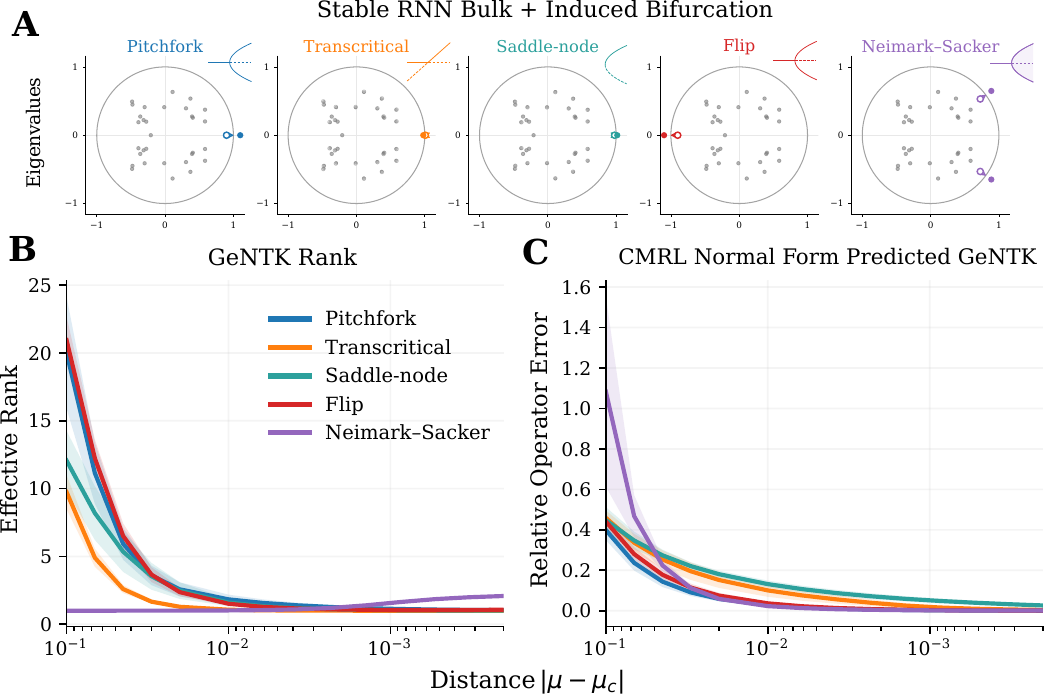}
    \caption{\textbf{CMRL accurately predicts the amplified low-rank learning geometry near all five bifurcations.}
For dense-coupled 32-dimensional tanh RNNs, we locate each bifurcation and evaluate the full parameter Jacobian with respect to all entries of \(W\) and \(b\).
\textbf{A} Eigenvalue spectra illustrating each controlled bifurcation embedded in a stable, densely coupled RNN bulk.
\textbf{B} GeNTK effective rank, approaching one for pitchfork, transcritical, saddle-node, and flip, and two for Neimark--Sacker.
\textbf{C} Relative Frobenius error of the CMRL-predicted GeNTK, which decreases sharply as the bifurcation is approached.
Shading denotes the mean \(\pm 1\) standard deviation across five seeds.}  
    \label{fig:valthm}
\end{figure}

\begin{theorem}[Center-Manifold Reduction of Learning]
\label{thm:cmfd}
Assume the center-manifold representation in
Equation~\eqref{eqn:cmrstate}, with
\(\|v_{\mathrm{bif}}\|=1\), and define
\[
\phi_{\mathrm{bif}}
    := d_\mu\alpha\otimes v_{\mathrm{bif}},
\qquad
J_{\mathrm{CMRL}}
    := \phi_{\mathrm{bif}}D_\theta\mu.
\]
Let \(\mathcal U_\rho\) be a local neighborhood of radius \(\rho\) around
the reference bifurcation, and suppose that throughout this neighborhood
\[
\|D_\alpha\varepsilon\|
    \leq \eta(\rho),
\qquad
\|D_\mu\varepsilon\|_F
    \leq M_\mu,
\]
\[
\|D_Rh\,D_\theta R\|_F
    \leq M_R,
\qquad
\|D_\theta\mu\|
    \geq m>0.
\]
Then
\[
\frac{\|D_\theta h-J_{\mathrm{CMRL}}\|_F}
     {\|J_{\mathrm{CMRL}}\|_F}
\leq
\eta(\rho)
+
\frac{M_\mu}{\|d_\mu\alpha\|_F}
+
\frac{M_R}{m\|d_\mu\alpha\|_F}.
\]
By tangency of the center manifold to the critical eigenspace,
\(\eta(\rho)\to0\) as \(\rho\to0\). Moreover, assuming
Proposition~\ref{prop:amp}, if the residual sensitivities remain bounded
while \(\|d_\mu\alpha\|_F\) becomes large, then
\[
\frac{\|D_\theta h-J_{\mathrm{CMRL}}\|_F}
     {\|J_{\mathrm{CMRL}}\|_F}
\leq
\eta(\rho)+o(1).
\]
Thus, in the joint local and amplified regime,
\[
D_\theta h
\approx
J_{\mathrm{CMRL}}
=
(d_\mu\alpha\otimes v_{\mathrm{bif}})D_\theta\mu.
\]
\end{theorem}

\paragraph{Numerical Validation}
We validate Theorem~\ref{thm:cmfd} in a minimal, controlled high-dimensional setting designed to isolate each of the five codimension-one bifurcations. 
For each case, we construct a 32-dimensional tanh RNN in which the relevant critical mode is densely mixed with a random stable bulk and weakly coupled to the remaining modes. 
We then numerically locate the resulting bifurcation and evaluate the full parameter-to-trajectory Jacobian \(D_\theta h\) with respect to all entries of \(W\) and \(b\), rather than restricting the parameterization to the planted critical direction. 
Figure~\ref{fig:valthm} shows that, as the bifurcation is approached, the GeNTK concentrates toward the predicted critical dimension---rank one for pitchfork, transcritical, saddle-node, and flip, and rank two for Neimark--Sacker---while the relative error of the CMRL-predicted GeNTK sharply decreases. 
Importantly, the CMRL field is constructed from the continuation-derived critical eigenspace and normal-form dynamics, without fitting to the measured Jacobian.
Appendix~\ref{app:controlled_validation} gives the full experimental construction and additionally shows amplification of both \(D_\theta h\) and the GeNTK, together with the direct relative error
\(\|D_\theta h-J_{\mathrm{CMRL}}\|_F/\|D_\theta h\|_F\).
Thus, this experiment is intended as a controlled numerical validation of the operator reduction in Theorem~\ref{thm:cmfd}, rather than a survey of generic random RNN bifurcations.

\paragraph{Testable Operator-Level Signatures of Bifurcations}
Theorem~\ref{thm:cmfd} yields three directly measurable signatures of a bifurcation-dominated learning channel.

\begin{corr}[Operator-Level Signatures of Codimension-One Bifurcations]
\label{corr:sig}
Suppose the approximation error in Theorem~\ref{thm:cmfd} is small. As the bifurcation channel dominates, the norm of \(D_\theta h\) is amplified and its effective rank approaches one. If the weak cross-talk condition also holds, its dominant state-space field approaches
\(\phi_{\mathrm{bif}}=d_\mu\alpha\otimes v_{\mathrm{bif}}\), which has Schmidt rank one across the sample--time and state-space partition. For a Neimark--Sacker bifurcation, the corresponding real state-space rank is at most two.
\end{corr}

\begin{figure}[t]
    \centering
    \includegraphics[width=\linewidth]{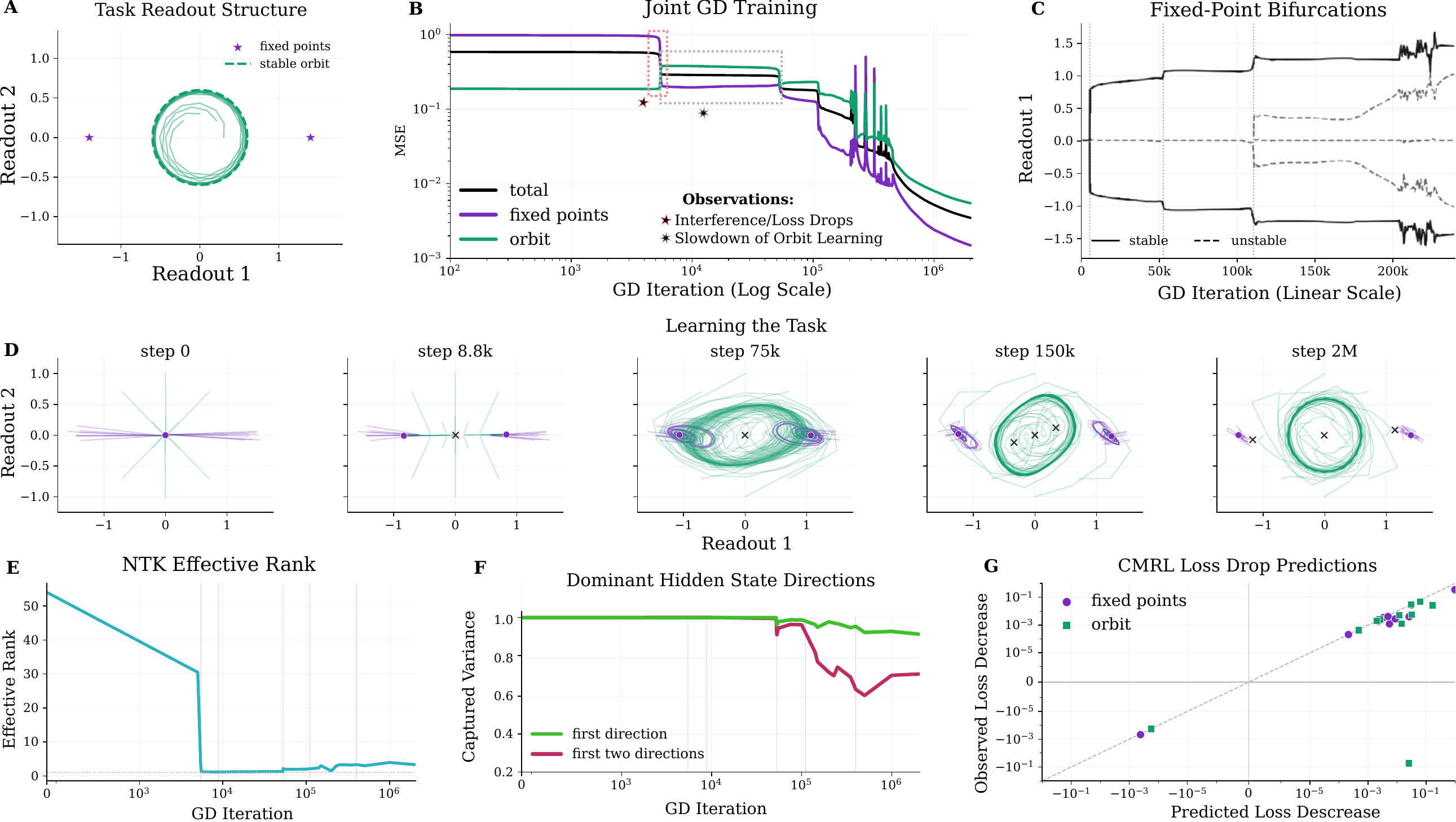}
    \caption{\textbf{Bifurcations reorganize learning in a joint fixed-point--orbit task.}
    \textbf{A} Initial conditions select either one of two stable fixed points or a stable orbit in a shared two-dimensional readout.
    \textbf{B} Joint-training losses exhibit an abrupt fixed-point loss drop, transient interference with the orbit objective, and a long period of slow orbit learning.
    \textbf{C} Stable and unstable fixed-point branches reveal the dynamical transitions underlying the early loss drops.
    \textbf{D} Readout trajectories across training show sequential formation and reorganization of the two motifs.
    \textbf{E} The state-space NTK effective rank collapses near the first acquisition event and remains low during much of subsequent learning.
    \textbf{F} The variance captured by the dominant hidden-state directions tracks the accompanying low-dimensional latent distortion.
    \textbf{G} CMRL predictions of the signed fixed-point and orbit loss changes compared with the observed changes. Most events lie near the identity line; large deviations indicate breakdown of the local rank-one approximation.}
    \label{fig:interference_main}
\end{figure}

These signatures can be estimated without forming the full operator using randomized trace estimation and matrix-free eigensolvers. They need not uniquely identify a strict bifurcation: slow-manifold changes or other critical transitions may produce similar amplification and spectral concentration. We therefore use them as diagnostics of critical low-dimensional learning geometry rather than as a stand-alone classifier of bifurcation type. Classical fixed- and slow-point analyses remain complementary~\citep{golub2018fixedpointfinder,driscoll2024flexible,sussillo2013opening}.

\section{Consequences for Learning}
\label{sec:consequences}

The CMRL approximation implies that, near a bifurcation, all task errors are filtered through the same amplified state-space field. This can produce a sudden loss drop for the task aligned with that field, transient interference with other tasks, distortion of the shared latent representation, and a prolonged reduction in the effective directions available for subsequent learning. We study these effects in one minimal autonomous RNN task.

\paragraph{Joint fixed-point--orbit task.}
The initial condition selects the required dynamical motif. Two lateral groups of initial conditions must converge to distinct stable fixed points, while a central group must approach a stable orbit, all through the same two-dimensional readout. Let \(L_{\mathrm{FP}}\) and \(L_{\mathrm{orb}}\) denote the corresponding trajectory-matching losses. We train the two objectives jointly throughout unless otherwise stated.

Figure~\ref{fig:interference_main} gives the full experimental overview. The network can ultimately represent both motifs, ruling out a simple capacity limitation, but learning is strongly sequential: the fixed-point motif is acquired near step \(8.8\mathrm{k}\), while the orbit loss remains elevated for far longer. We first state the local prediction, then separate its four main consequences.

\begin{corr}[Bifurcation-Induced Tunnel Vision]
\label{cor:bif_interference}
Assume Theorem~\ref{thm:cmfd} and let
\(\phi_{\mathrm{bif}}=d_\mu\alpha\otimes v_{\mathrm{bif}}\).
For \(L=\sum_{m=1}^M L_m\), define
\(a_m:=\inner{\nabla_h L_m}{\phi_{\mathrm{bif}}}\),
\(a:=\sum_m a_m\), and
\(c:=\norm{\nabla_\theta\mu}^2\).
Then, to leading order,
\[
    \dot h\approx-ca\,\phi_{\mathrm{bif}},
    \qquad
    \frac{dL_m}{ds}\approx-ca_m a,
    \qquad
    \NTK\approx c\,\phi_{\mathrm{bif}}\otimes\phi_{\mathrm{bif}}.
\]
Hence subtask \(m\) improves when \(a_m a>0\), is locally invisible when
\(a_m=0\), and worsens when \(a_m a<0\). Learning and representational
change are simultaneously concentrated along the same bifurcation field.
\end{corr}

\subsection{Sudden Loss Drops}
\label{subsec:loss_drops}

The first fixed-point bifurcations coincide with an abrupt decrease in
\(L_{\mathrm{FP}}\), a collapse in NTK effective rank, and a qualitative
reorganization of the readout trajectories
(Figure~\ref{fig:interference_main}B--E).
Under Corollary~\ref{cor:bif_interference}, this occurs when the fixed-point
error is aligned with the amplified bifurcation field, so that
\(a_{\mathrm{FP}}a>0\). The large norm of
\(\phi_{\mathrm{bif}}\) then concentrates the GD update along precisely the
state-space direction required to form the new motif. Thus, the sudden loss
drop is not simply learning becoming uniformly faster: it reflects the
creation of a highly selective learning channel.

Panel~G tests this prediction directly by comparing the CMRL-predicted
fixed-point loss changes with the observed changes across the main events.
The close agreement for most checkpoints supports the predicted connection
between the bifurcation field and the abrupt loss decrease.

\subsection{Sub-Task Interference}
\label{subsec:interference}

The same bifurcation can simultaneously harm a competing objective. Near the
first fixed-point acquisition event, the fixed-point loss drops while the
orbit loss increases (Figure~\ref{fig:interference_main}B). Corollary~
\ref{cor:bif_interference} predicts this behavior when
\(a_{\mathrm{orb}}a<0\): the total gradient is dominated by a bifurcation
field that helps the fixed-point objective but moves the orbit trajectories
in the wrong direction.

This is the central ``tunnel vision'' effect. The bifurcation does not merely
produce a large gradient; it forces multiple task errors through nearly the
same state-space direction. Aligned errors improve together, orthogonal
errors are temporarily ignored, and opposing errors worsen. Panel~G shows
that the projected fixed-point and orbit errors generally predict the signs
of these changes. The isolated large deviations are also informative: they
identify regimes where residual channels or center--residual cross-talk are
no longer negligible.

\subsection{Latent Reorganization}
\label{subsec:latent_distortion}

The same low-rank event is accompanied by a reorganization of the shared latent representation. The rank collapse in Figure~\ref{fig:interference_main}E shows that learning becomes concentrated into very few state-space directions, while panels D and F show corresponding changes in the trajectories and hidden-state variance. This is consistent with a self-referential effect in which an amplified state-space mode changes the sampled trajectories and thereby alters subsequent parameter sensitivity~\citep{hazelden2026globalempiricalntkselfreferential}. We do not require exact alignment between the dominant hidden-state variance direction and \(\phi_{\mathrm{bif}}\) for the local loss-change result above.

\subsection{Low-Rank Natural-Gradient Intervention}
\label{subsec:lowrank_ng}

\paragraph{Motivation.}
The same low-rank concentration that produces tunnel vision suggests that the
resulting stiffness may be addressed without approximating a full natural gradient. Near a bifurcation, a small critical subspace can dominate the learning geometry, motivating a correction restricted to a few leading directions. Related low-rank natural-gradient methods have been developed in
other settings~\citep{mishkin2018slang,yang2022seng,yang2020sketchy}. Here, we briefly test such a correction, running SGD on the remaining parameter directions while reconditioning a small number of leading trajectory-residual Gauss--Newton modes.
\paragraph{Low-rank correction.}
We therefore correct only the leading modes of the trajectory-residual
Gauss--Newton operator. The leading directions are computed matrix-free and
reconditioned with a damped inverse, while the update on the orthogonal
complement remains ordinary SGD. In the experiment below, these modes are
recomputed every update using the full trajectory residual. Implementation
details are given in Appendix~\ref{app:lowrank_ng}.

\begin{figure}[t]
    \centering
    \includegraphics[width=0.7\linewidth]{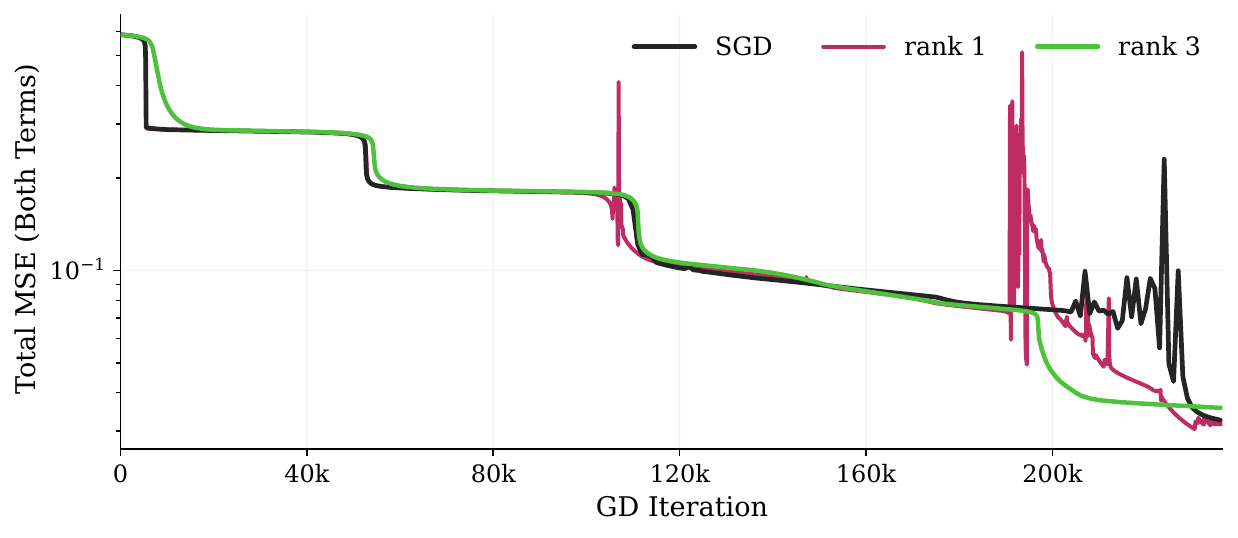}
    \caption{\textbf{Low-rank natural-gradient intervention in the setting of Figure~\ref{fig:interference_main}.} A rank-three correction removes the large loss excursions over the common training horizon, whereas a rank-one correction does not.}
    \label{fig:lowrankng}
\end{figure}

\paragraph{Results.}
Figure~\ref{fig:lowrankng} compares SGD with rank-one and rank-three
corrections on the joint fixed-point--orbit task. Correcting only one mode
does not remove the large loss excursions, whereas the rank-three correction
gives a smooth, monotonically decreasing loss over the common training
horizon. This is consistent with stabilization by reconditioning a small
critical subspace. We do not, however, identify the three corrected modes
individually with the one-dimensional fixed-point channel and the
two-dimensional real oscillatory channel, so we treat this intervention as
supportive rather than a direct identification of those modes.

\section{Operator Signatures in an Input-Driven Multitask LeakyRNN}
\label{sec:multitask_gru}
\paragraph{Motivation.}
Multitask recurrent networks can solve related cognitive tasks by reusing dynamical motifs, including attractors, decision boundaries, and rotations; in particular, tasks requiring memory of a continuous variable can reuse a common ring attractor~\citep{driscoll2024flexible}. Establishing such reuse classically requires detailed dynamical analysis tailored to each task and input epoch. We therefore ask whether the matrix-free GeNTK diagnostics implied by Corollary~\ref{corr:sig} reflect task-dependent changes in trajectory sensitivity, including transitions induced by inputs rather than by parameter changes.

\begin{figure}[t]
    \centering
    \includegraphics[width=\linewidth]{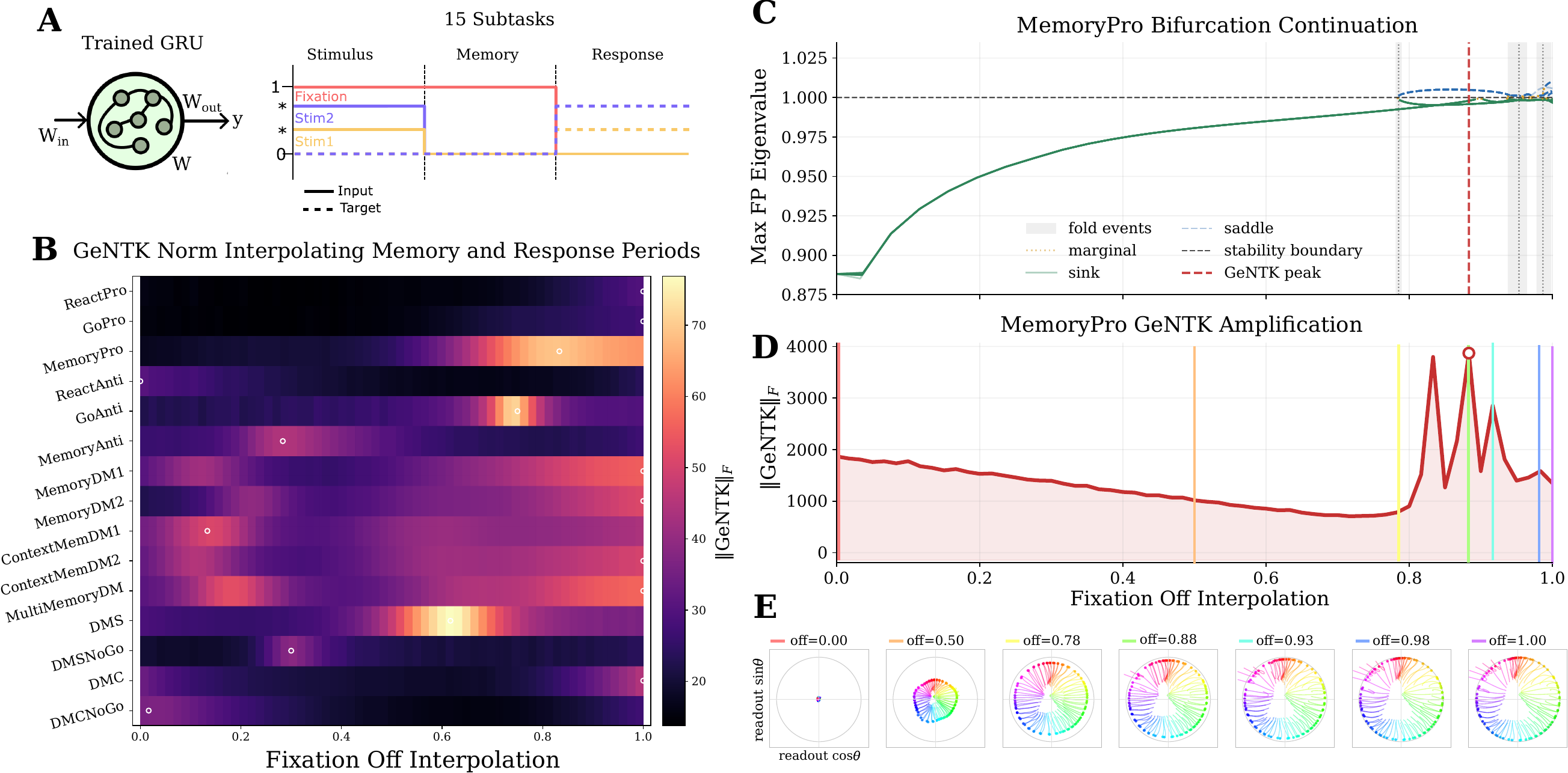}
    \caption{\textbf{Operator signatures across 15 tasks in a multitask LeakyRNN.}
    \textbf{A} Simplified task schematic, adapted from~\citet{driscoll2024flexible}.
    \textbf{B} GeNTK amplification for each task as the fixation-off input is interpolated from \(0\) (response) to \(1\) (memory), revealing task-dependent transitions between response and memory dynamics.
    \textbf{C--E} MemoryPro case study.
    \textbf{C} Fixed-point continuation shows a transition from a single stable fixed point at fixation \(0\) to a family of marginal fixed points near fixation \(1\), supporting stimulus retention during memory.
    \textbf{D} GeNTK amplification peaks in the same interpolation range as these continuation-detected changes.
    \textbf{E} Example trajectories show collapse to a fixed point at fixation \(0\) and formation of a ring attractor near fixation \(1\).}
    \label{fig:mutlitask}
\end{figure}

\paragraph{Experimental setup.}
We analyze an input-driven LeakyRNN trained jointly on 15 tasks from this multitask family. For each task, we interpolate the fixation-off input from \(0\), corresponding to the response-period dynamics, to \(1\), corresponding to the memory-period dynamics, while holding the task rule and stimulus condition fixed. During each constant-input epoch this defines a frozen-input dynamical system, and we estimate its GeNTK amplification matrix-free across the interpolation. For MemoryPro, we additionally perform fixed-point continuation and simulate trajectories from multiple initial conditions, providing an independent dynamical-systems validation of the operator diagnostic. Additional setup details are given in Appendix~\ref{app:input_gru}.

\paragraph{Results.}
Figure~\ref{fig:mutlitask}B reveals structured, task-dependent amplification bands between the response and memory regimes across the 15 tasks. In MemoryPro, the largest GeNTK peaks occur in the same interpolation range where continuation detects marginal and stability-changing fixed-point branches (Figure~\ref{fig:mutlitask}C--D). The corresponding trajectories change from collapse toward a single stable fixed point at fixation \(0\) to a ring-like family that retains the stimulus near fixation \(1\) (Figure~\ref{fig:mutlitask}E), consistent with the memory motif identified in the original multitask analysis. We view this as evidence that the matrix-free operator diagnostic can localize input-dependent changes in learning geometry; establishing motif reuse from the GeNTK alone would require additional task-by-task dynamical analysis.

\section{Discussion}

\paragraph{Takeaways.}
Bifurcations affect learning because the parameter-to-state Jacobian can become dominated by one amplified channel. CMRL identifies this channel with a normal-form sensitivity and a bifurcation-relevant state direction, turning a high-dimensional local learning problem into an explicit low-rank approximation. The resulting geometry is not simply fast: it is selective. Aligned error signals are learned rapidly, orthogonal signals are delayed, and anti-aligned signals can be harmed.

The joint fixed-point--orbit task makes these consequences testable in one controlled setting. The key empirical question is whether the bifurcation field predicts the signs of subtask loss changes while spectral concentration coincides with low-dimensional latent reorganization. The agreement is strongest near isolated events, while deviations provide a direct indication that residual channels or center--residual cross-talk are no longer negligible.

The present theory is local and finite-horizon. It assumes that one bifurcation channel dominates and that residual cross-talk is sufficiently weak for the normal-form field to remain predictive. Closely spaced bifurcations, strongly coupled center and transverse dynamics, and input-driven transitions may violate these assumptions. These limitations also suggest the main practical direction: use randomized operator diagnostics to detect when the approximation is valid, rather than assuming that every loss jump or rank collapse corresponds to an isolated classical bifurcation.

\acks{We thank Laura Driscoll for support and discussion of this work. In addition, we thank Michael Buice, David Bell, Hanson Mo, Saba Heravi, Julian Brandon, Owen Marschall and Blake Bordelon for motivating discussion related to aspects of the work. Finally, we acknowledge Alexander Hsu for suggesting the low-rank natural-gradient optimizer as a possible avenue for overcoming blowup of a few GeNTK modes.}

\bibliography{refs}

\newpage
\clearpage
\appendix
\section{Student--Teacher RNN Setup}
\label{app:student_teacher}

Figure~\ref{fig:rnn} uses a student--teacher trajectory-matching task. The student and teacher receive the same initial conditions, and the student is trained to match the teacher's two-dimensional readout trajectories. The example is used to visualize a learned pitchfork event and the accompanying change in the empirical parameter-to-state geometry; the controlled five-bifurcation experiment in Appendix~\ref{app:controlled_validation} is the quantitative test of the CMRL approximation.

\section{Input-Driven 15-Task LeakyRNN}
\label{app:input_gru}

We analyze a LeakyRNN trained jointly on the 15 tasks shown in Figure~\ref{fig:mutlitask}, using the input--output organization of~\citet{driscoll2024flexible}. For each task, the rule and stimulus are held fixed while the fixation-off channel is interpolated between the response-period value and the memory-period value. At each interpolation value, we collect the hidden-state trajectory and estimate the GeNTK amplification using matrix-free Jacobian products.

For MemoryPro, we additionally freeze the interpolated input during the memory epoch, solve for fixed points from multiple initial seeds, and continue the resulting branches across the interpolation. Fixed points are classified from the spectrum of the frozen-input recurrent Jacobian. We compare the resulting bifurcation locations with the GeNTK peaks and visualize trajectories from multiple initial conditions in the two-dimensional readout. 

\section{Low-Rank Natural-Gradient Details}
\label{app:lowrank_ng}

Let \(J=D_\theta h\), and estimate the dominant eigenpair \((u_1,\lambda_1)\) of \(\NTK=JJ^\ast\) by power iteration using matrix-free applications of \(J\) and \(J^\ast\). The corresponding normalized parameter-space Fisher direction is \(v_1=J^\ast u_1/\sqrt{\lambda_1}\). The rank-one correction applies SGD on the orthogonal complement of \(v_1\) and rescales the component along \(v_1\) by the inverse damped eigenvalue. This requires only a small number of additional Jacobian products when the spectral gap is large.

\section{Numerical Continuation and CMRL Extraction}
\label{app:continuation}

For the autonomous RNN experiments, numerical continuation follows fixed-point branches as the control parameter varies and detects codimension-one events through crossings of the unit circle. At each event, the critical state-space direction and unfolding coordinate define \(\phi_{\mathrm{bif}}\) and \(J_{\mathrm{CMRL}}\) in Algorithm~\ref{alg:cmrl}. Operator quantities are evaluated matrix-free using Jacobian--vector and vector--Jacobian products.

\section{Proof of Theorem~\ref{thm:cmfd}}
\label{app:cmrl_proof}

We take \(\bar h\) and \(v_{\mathrm{bif}}\) at the reference bifurcation,
absorbing any remaining parameter dependence into \(\varepsilon\).
In the local coordinates
\(\theta\leftrightarrow(\mu,R)\), the chain rule gives
\[
D_\theta h
=
D_\mu h\,D_\theta\mu
+
D_Rh\,D_\theta R.
\]
Differentiating Equation~\eqref{eqn:cmrstate} with respect to \(\mu\)
gives
\[
D_\mu h
=
d_\mu\alpha\otimes v_{\mathrm{bif}}
+
D_\alpha\varepsilon[d_\mu\alpha]
+
D_\mu\varepsilon.
\]
Therefore,
\[
D_\theta h
=
J_{\mathrm{CMRL}}
+
D_\alpha\varepsilon[d_\mu\alpha]D_\theta\mu
+
D_\mu\varepsilon D_\theta\mu
+
D_Rh\,D_\theta R,
\]
and consequently
\[
\|D_\theta h-J_{\mathrm{CMRL}}\|_F
\leq
\|D_\alpha\varepsilon[d_\mu\alpha]D_\theta\mu\|_F
+
\|D_\mu\varepsilon D_\theta\mu\|_F
+
\|D_Rh\,D_\theta R\|_F.
\]

Since \(\|v_{\mathrm{bif}}\|=1\),
\[
\|J_{\mathrm{CMRL}}\|_F
=
\|d_\mu\alpha\|_F\|D_\theta\mu\|.
\]
The center--residual cross-talk term therefore satisfies
\[
\frac{
\|D_\alpha\varepsilon[d_\mu\alpha]D_\theta\mu\|_F
}{
\|J_{\mathrm{CMRL}}\|_F
}
\leq
\|D_\alpha\varepsilon\|
\leq
\eta(\rho).
\]
Similarly,
\[
\frac{
\|D_\mu\varepsilon D_\theta\mu\|_F
}{
\|J_{\mathrm{CMRL}}\|_F
}
\leq
\frac{M_\mu}{\|d_\mu\alpha\|_F},
\]
and, using \(\|D_\theta\mu\|\geq m\),
\[
\frac{
\|D_Rh\,D_\theta R\|_F
}{
\|J_{\mathrm{CMRL}}\|_F
}
\leq
\frac{M_R}{m\|d_\mu\alpha\|_F}.
\]
Combining the three bounds gives
\[
\frac{\|D_\theta h-J_{\mathrm{CMRL}}\|_F}
     {\|J_{\mathrm{CMRL}}\|_F}
\leq
\eta(\rho)
+
\frac{M_\mu}{\|d_\mu\alpha\|_F}
+
\frac{M_R}{m\|d_\mu\alpha\|_F}.
\]

By tangency of the center manifold to the critical eigenspace,
\(\eta(\rho)\to0\) as the neighborhood shrinks. Assuming
Proposition~\ref{prop:amp}, normal-form amplification suppresses the
remaining bounded residual terms, completing the proof.

\section{Controlled High-Dimensional Validation of CMRL}
\label{app:controlled_validation}

Figure~\ref{fig:valthm} provides a minimal controlled test of
Theorem~\ref{thm:cmfd} in high-dimensional nonlinear RNNs.
The goal is not to sample generic bifurcations from random networks, but to
construct each of the five canonical events in a setting where the critical
mode is embedded in and coupled to a genuinely high-dimensional dynamical
system, and then test whether CMRL predicts the full parameter-to-trajectory
operator.

\paragraph{Dense-coupled RNN construction.}
For each bifurcation, we use a 32-dimensional autonomous tanh RNN,
\[
    h_{t+1}=\tanh(W(\mu)h_t+b(\mu)).
\]
The desired critical linearization is embedded into a random stable bulk
through
\[
    A(\mu)
    =
    Q\left[B_{\mathrm{crit}}(\mu)\oplus S\right]Q^T
    +\epsilon G,
\]
where \(Q\) is a dense random orthogonal matrix, \(S\) is a stable random
bulk with eigenvalues strictly inside the unit circle, and \(G\) introduces
dense coupling between the critical and residual directions. We use
\(\epsilon=0.05\) in the main experiment. Thus, although the construction
allows us to control which bifurcation occurs, the critical direction is
neither an isolated neuron nor an invariant coordinate block.

The critical block is one-dimensional for pitchfork, transcritical,
saddle-node, and flip, and two-dimensional for Neimark--Sacker. Dense
coupling shifts the event away from its nominal location, so in each
realization we numerically locate the actual bifurcation value \(\mu_c\)
from the relevant multiplier crossing: \(+1\) for pitchfork,
transcritical, and saddle-node, \(-1\) for flip, and the unit circle for
the complex-conjugate Neimark--Sacker pair.

Pitchfork, flip, and Neimark--Sacker are constructed around the origin with
zero bias. For the transcritical and saddle-node cases, we use a nonzero
reference fixed point so that the tanh nonlinearity provides the required
quadratic term. The transcritical construction preserves one fixed-point
branch while a second branch crosses and exchanges stability. The
saddle-node is unfolded through a bias perturbation with nonzero projection
onto the critical left eigendirection, producing the expected one-sided
stable--unstable fixed-point pair.

\paragraph{Full parameter-to-trajectory Jacobian.}
Crucially, the differentiated parameter vector contains every recurrent
weight and bias,
\[
    \theta=(\operatorname{vec}W,b)\in\R^{1056}.
\]
We therefore evaluate the full empirical Jacobian
\[
    J:=D_\theta h,
\]
rather than differentiating only with respect to \(\mu\) or the planted
critical block. Trajectories are propagated for \(T=1024\) recurrent steps,
with the final six steps retained for two sampled trajectories. This gives
\[
    J\in\R^{384\times1056},
    \qquad
    \Theta:=JJ^T\in\R^{384\times384}.
\]
For this controlled experiment these operators are evaluated directly,
rather than by randomized sketching.

\begin{figure}[t]
    \centering
    \includegraphics[width=\linewidth]{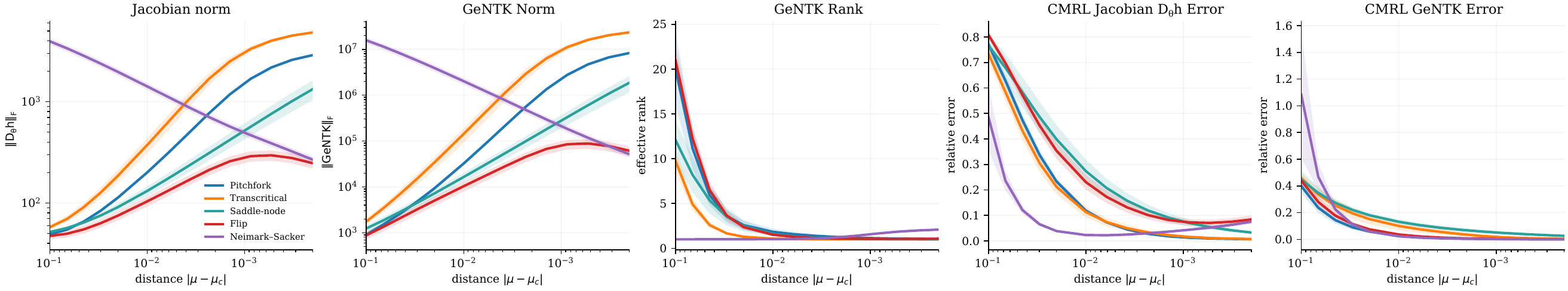}
    \caption{\textbf{Full operator validation of CMRL in dense-coupled
    32-dimensional tanh RNNs.}
    From left to right, we show the Frobenius norm of the full
    parameter-to-trajectory Jacobian \(J=D_\theta h\), the Frobenius norm
    of the induced GeNTK \(\Theta=JJ^T\), GeNTK effective rank, the direct
    relative Jacobian error
    \(\|J-J_{\mathrm{CMRL}}\|_F/\|J\|_F\), and the relative GeNTK error
    \(\|\Theta-\Theta_{\mathrm{CMRL}}\|_F/\|\Theta\|_F\).
    Curves show the mean across five independent random seeds and shading
    denotes one standard deviation. The scalar bifurcations concentrate
    toward a one-dimensional critical channel, while Neimark--Sacker
    approaches the two-dimensional real critical subspace.}
    \label{fig:valthm_full}
\end{figure}

For each bifurcation we evaluate 14 distances from the numerically located
event,
\[
    |\mu-\mu_c|\in[2\times10^{-4},10^{-1}],
\]
and repeat the construction across five independent random seeds. Each seed
changes the bulk, dense rotation, coupling, and trajectory initialization.
The curves in Figures~\ref{fig:valthm} and~\ref{fig:valthm_full} show the
mean across seeds, with shading denoting one standard deviation.

\paragraph{CMRL prediction.}
At each corrected event \(\mu_c\), we extract the critical left and right
eigenspaces from the recurrent state Jacobian. The corresponding real
critical dimension is one for pitchfork, transcritical, saddle-node, and
flip, and two for Neimark--Sacker. We then propagate the reduced tangent
dynamics within this critical eigenspace, including the projected parameter
forcing at each recurrent step, to construct \(J_{\mathrm{CMRL}}\).
No singular vectors of the measured full Jacobian are used to fit this
approximation. The corresponding predicted GeNTK is
\[
    \Theta_{\mathrm{CMRL}}
    =
    J_{\mathrm{CMRL}}J_{\mathrm{CMRL}}^T.
\]

\paragraph{Results.}
Figure~\ref{fig:valthm_full} shows the quantities underlying the compressed
main-text validation. For the four scalar bifurcations, approaching the
event produces strong amplification of the Jacobian and GeNTK together
with collapse of the GeNTK effective rank toward one. At the closest
sampled point, the mean effective ranks are approximately \(1.04\),
\(1.00\), \(1.01\), and \(1.07\) for pitchfork, transcritical,
saddle-node, and flip, respectively. The direct CMRL Jacobian error also
falls sharply near the event, reaching approximately \(0.7\%\), \(0.8\%\),
\(3.3\%\), and \(8.4\%\) in these four cases. The induced GeNTK is even
more accurately predicted, with relative errors of only a few percent or
less.

Neimark--Sacker behaves slightly differently because the post-bifurcation
trajectory lies on an invariant circle. Its long-horizon phase sensitivity
can already dominate the empirical Jacobian away from the event, so its
effective-rank and norm curves need not exhibit the same monotone
high-rank-to-low-rank transition as the scalar cases. Nevertheless, as the
event is approached, the CMRL error falls sharply and the GeNTK approaches
the predicted two-dimensional real critical subspace; at the closest
sampled point its effective rank is approximately \(2.08\), with direct
Jacobian error approximately \(7.5\%\) and GeNTK error below \(1\%\).
Thus, the robust prediction across all five cases is concentration into the
continuation-derived critical subspace and improving CMRL operator accuracy
near the event, rather than an identical monotone norm profile for every
trajectory protocol.

This experiment is deliberately controlled. It tests the quantitative
operator approximation while keeping the bifurcation type known and
isolated, but does not claim that arbitrary random RNNs generically exhibit
the same clean separation. The learned RNN and input-driven LeakyRNN experiments
in the main text test complementary consequences and observable signatures
in less controlled settings.

\section{Normal-Form Sensitivity}
\label{app:normal_form_sensitivity}

Iterating the sensitivity recurrence in
Proposition~\ref{prop:amp} gives
\[
(d_\mu\alpha)_t
=
\sum_{k=0}^{t-1}
\partial_\mu g(\alpha_k,\mu)
\prod_{r=k+1}^{t-1}
\partial_\alpha g(\alpha_r,\mu).
\]
Thus each parameter perturbation is propagated by the subsequent local
multipliers. Near a bifurcation these products decay slowly, producing the
amplification shown in Figure~\ref{fig:normformamp}.

For the linear critical case
\[
\alpha_{t+1}=(1+\mu)\alpha_t,
\]
we have exactly
\[
(d_\mu\alpha)_t=t(1+\mu)^{t-1}\alpha_0.
\]
At \(\mu=0\),
\[
\|d_\mu\alpha\|_2^2
=
\alpha_0^2\sum_{t=1}^{T-1}t^2
=
O(T^3),
\]
and therefore
\[
\|d_\mu\alpha\|_2=O(T^{3/2}).
\]
For the nonlinear normal forms, motion onto a stable branch or orbit limits
this growth after the bifurcation.

\end{document}